\documentclass[10pt,twocolumn,letterpaper]{article}

\usepackage{cvpr}
\usepackage{times}
\usepackage{epsfig}
\usepackage{graphicx}
\usepackage{amsmath}
\usepackage{amssymb}
\usepackage{bbm}
\usepackage{enumitem, url}
\usepackage{comment}

\usepackage{color}

\newcommand{\task}{\mbox{Interactive Question Answering}} 
\newcommand{\taskshort}{\mbox{\sc IQA}}
\newcommand{\dataset}{\mbox{Interactive Question Answering Dataset}}
\newcommand{\datasetshort}{\mbox{\sc iquad v1}}
\newcommand{\model}{Hierarchical Interactive Memory Network}
\newcommand{\modelshort}{\mbox{\sc himn}}
\newcommand{\gru}{Egocentric Spatial GRU}
\newcommand{\grushort}{\mbox{esGRU}}

\usepackage[pagebackref=true,breaklinks=true,letterpaper=true,colorlinks,bookmarks=false]{hyperref}

\cvprfinalcopy 

\def\cvprPaperID{1352} 

\ifcvprfinal\fi
\begin{document}

\title{IQA: Visual Question Answering in Interactive Environments}

\author{Daniel Gordon$^1$ \quad Aniruddha Kembhavi$^2$ \quad Mohammad Rastegari$^{2,4}$ \\
Joseph Redmon$^1$ \quad Dieter Fox$^{1,3}$ \quad Ali Farhadi$^{1,2}$\\
{\normalsize $^1$Paul G. Allen School of Computer Science, University of Washington} \\
{\normalsize $^2$Allen Institute for Artificial Intelligence} \\
{\normalsize $^3$Nvidia} \quad {\normalsize $^4$Xnor.ai}
}
\maketitle
\begin{abstract}

We introduce \task\ (\taskshort), the task of answering questions that require an autonomous agent to interact with a dynamic visual environment. \taskshort\ presents the agent with a scene and a question, like: ``Are there any apples in the fridge?'' The agent must navigate around the scene, acquire visual understanding of scene elements, interact with objects (e.g. open refrigerators) and plan for a series of actions conditioned on the question. Popular reinforcement learning approaches with a single controller perform poorly on \taskshort\ owing to the large and diverse state space. We propose the \model\ (\modelshort), consisting of a factorized set of controllers, allowing the  system to operate at multiple levels of temporal abstraction.
To evaluate \modelshort, we introduce \datasetshort, a new dataset built upon AI2-THOR~\cite{thor}, a simulated photo-realistic environment of configurable indoor scenes with interactive objects.\footnote{For the full dataset and code release, visit \url{https://github.com/danielgordon10/thor-iqa-cvpr-2018}.}
\datasetshort\ has 75,000 questions, each paired with a unique scene configuration. Our experiments show that our proposed model outperforms popular single controller based methods on \datasetshort. For sample questions and results, please view our video: \url{https://youtu.be/pXd3C-1jr98}. 

\end{abstract}

\vspace{-5mm}
\section{Introduction}

A longstanding goal of the artificial intelligence community has been to create agents that can perform manual tasks in the real world and can communicate with humans via natural language. For instance, a household robot might be posed the following questions: \emph{Do we need to buy more milk?} which would require it to navigate to the kitchen, open the fridge and check to see if there is sufficient milk in the milk jug, or \emph{How many boxes of cookies do we have?} which would require the agent to navigate to the cabinets, open several of them and count the number of cookie boxes. Towards this goal, Visual Question Answering (VQA), the problem of answering questions about visual content, has received significant attention from the computer vision and natural language processing communities. While there has been a lot of progress on VQA, research by and large focuses on answering questions passively about visual content, i.e. without the ability to interact with the environment generating the content. An agent that is only able to answer questions passively is limited in its capacity to aid humans in their tasks.

\begin{figure}[t]
\begin{center}
\includegraphics[width=1.0\linewidth]{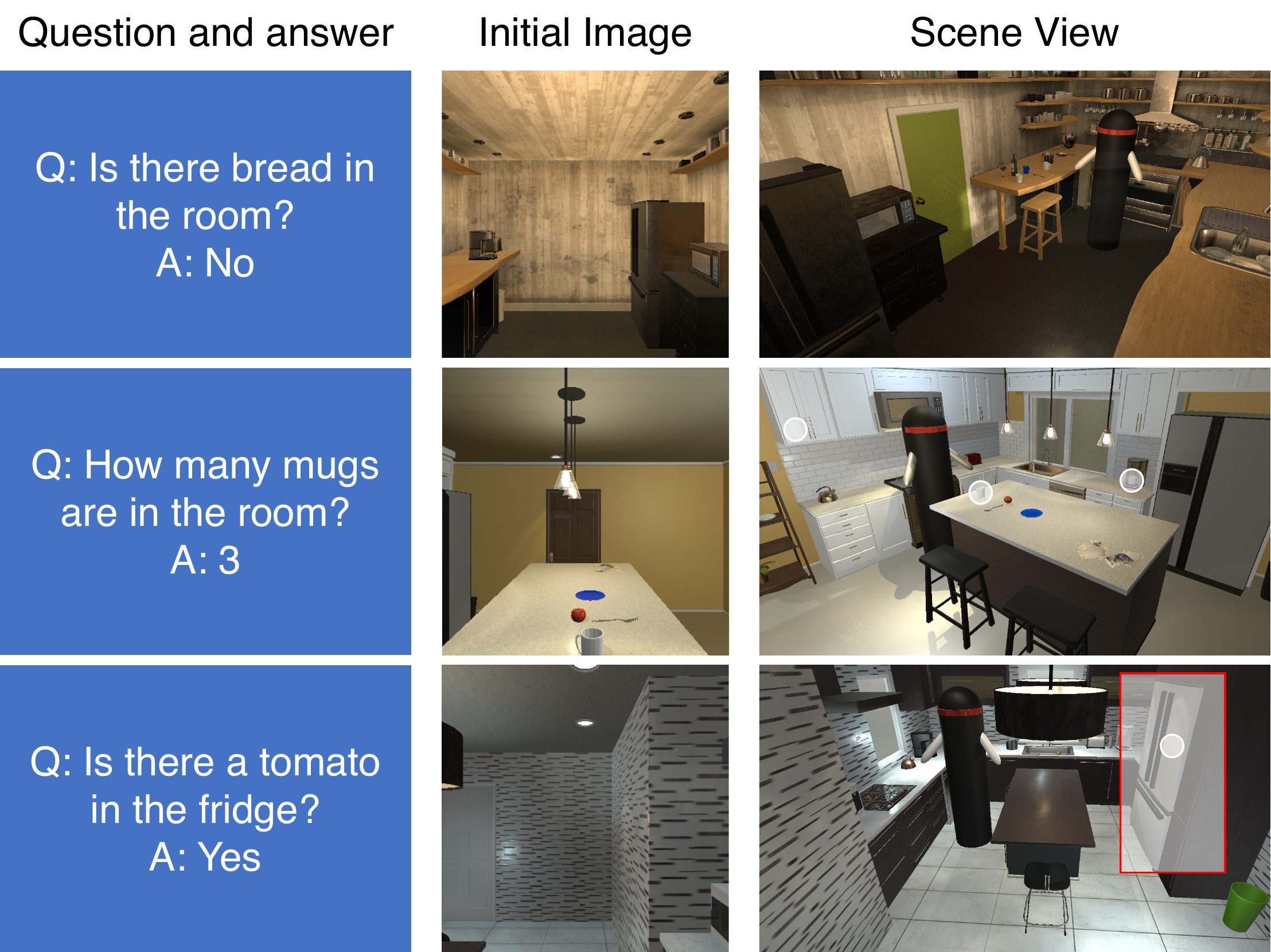}
\caption{Samples from \datasetshort: Each row shows a question paired with the agent's initial view and a scene view of the environment (which is not provided to the agent). In the scene view, the agent is shown in black, and the locations of the objects of interest for each question are outlined. Note that none of the questions can be answered accurately given only the initial image.} 
\vspace{-7mm}
\label{fig:teaser}
\end{center}
\end{figure}

\begin{figure}[t]
\begin{center}
\includegraphics[width=1.0\linewidth]{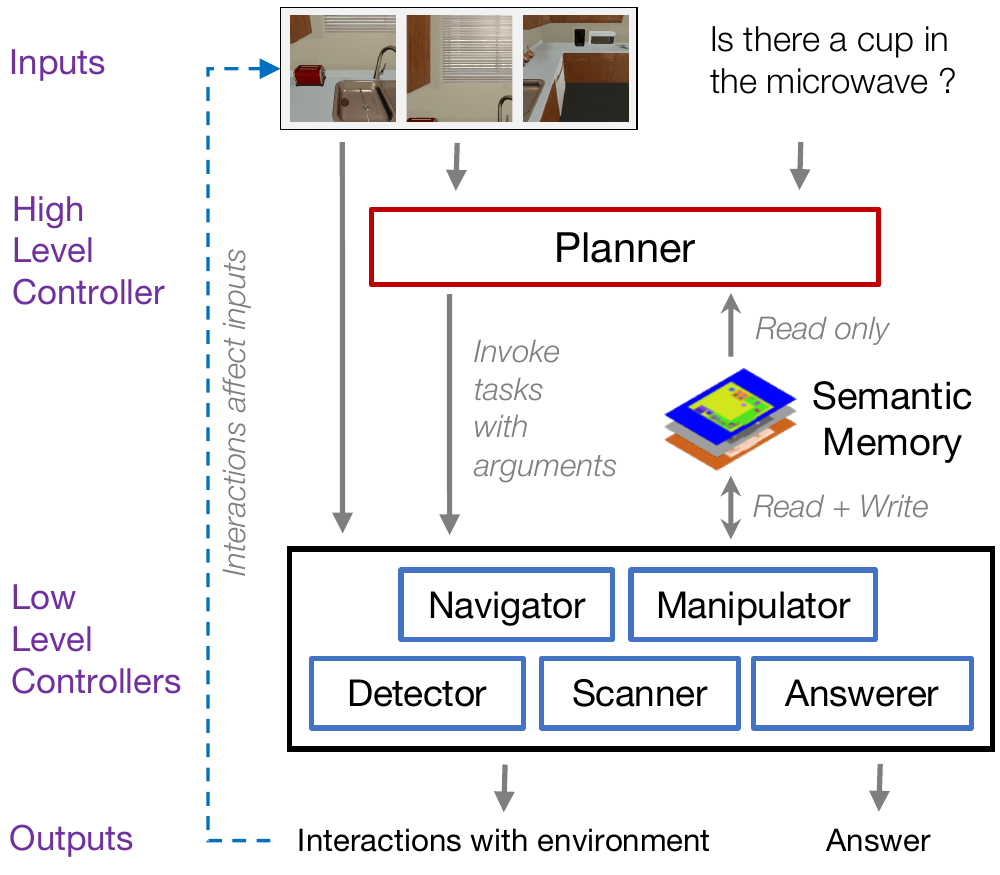}
\caption{An overview of the \model\ (\modelshort)} 
\vspace{-10mm}
\label{fig:overview}
\end{center}
\end{figure}

We introduce \textbf{\task}\ (\taskshort), the task of answering questions that require the agent to interact with a dynamic environment. \taskshort\ poses several key challenges in addition to the ones posed by VQA. \textbf{First}, the agent must be able to navigate through the environment. \textbf{Second}, it must acquire an understanding of its environment including objects, actions, and affordances. \textbf{Third}, the agent must be able to interact with objects in the environment (such as opening the microwave, picking up books, etc.). \textbf{Fourth}, the agent must be able to plan and execute a series of actions in the environment conditioned on the questions asked of it. 

To address these challenges, we propose \modelshort\ (\model). Figure~\ref{fig:overview} provides an overview of \modelshort. Akin to past works on hierarchical reinforcement learning, \modelshort\ is factorized into a hierarchy of controllers, allowing the system to operate, learn, and reason across multiple time scales while simultaneously reducing the complexity of each individual subtask. A high level controller, referred to as the \emph{Planner} chooses the task to be performed (for example, navigation / manipulation / answering / etc.) and generates a command for the chosen task. Tasks specified by the \emph{Planner} are executed by a set of low level controllers (\emph{Navigator}, \emph{Manipulator}, \emph{Detector}, \emph{Scanner} and \emph{Answerer}) which return control to the \emph{Planner} when a task termination state is reached.
Since these subtasks are fairly independent, we can pretrain each controller independently, while assuming oracle versions of the remaining controllers. Our experiments show that this factorization enables higher accuracy and generalization to unseen environments.

Several question types require the agent to remember where it has been and what it has seen. For example, \textit{How many pillows are in this house?} requires an agent to navigate around the rooms, open closets and keep track of the number of pillows it encounters. For sufficiently complex spaces, the agent needs to hold this information in memory for a long time. This motivates the need for an explicit external memory representation that is filled by the agent as it interacts with its environment. This memory must be both spatial and semantic so it can represent \textit{what} is \textit{where}. We propose a new recurrent layer formulation: \gru\ (\grushort) to represent this memory (Sec~\ref{sec:spatial_memory}). 

Training and evaluating interactive agents in the real world is currently prohibitive from the standpoint of operating costs, scale and research reproducibility. A far more viable alternative is to train and evaluate such agents in realistic simulated environments. Towards this end, we present the \dataset\ (\datasetshort) built upon AI2-THOR~\cite{thor}, a photo-realistic customizable simulation environment for indoor scenes integrated with the Unity~\cite{unity} physics engine. \datasetshort\ consists of over 75,000 multiple choice questions, each question accompanied by a unique scene configuration. 

We evaluate \modelshort\ on \datasetshort\ using a question answering accuracy metric and show that it outperforms a baseline based on a common architecture for reinforcement learning used in past work. We evaluate in both familiar and unfamiliar environments to show that our semantic model generalizes well across scenes.

In summary, our contributions include: (a) proposing \task, the task of answering questions that require the agent to interact with a dynamic environment, (b) presenting the \model, a question answering model factorized into a high level \emph{Planner}, a set of low level controllers and a rich semantic spatial memory, (c) the \gru, a new recurrent layer to represent this memory and (d) a new dataset \datasetshort\ towards the task of \taskshort.

\section{Related Work}

\noindent \textbf{Visual Question Answering (VQA):}
VQA has seen significant progress over the past few years, owing to the design of deep architectures suited for this task and the creation of large VQA datasets to train these models~\cite{Wu2016VisualQA}. These include datasets of natural images~\cite{agrawal2017vqa, Krishna2016VisualGC, malinowski2014multi, Wang2017FVQAFV}, synthetic images~\cite{agrawal2017vqa, Andreas2016NeuralMN, Johnson2016CLEVRAD,  Kahou2017FigureQAAA, Kembhavi2016ADI}, natural videos~\cite{Jang2017TGIFQATS, Tapaswi2016MovieQAUS}, synthetic videos~\cite{Kim2017DeepStoryVS, Mun2016MarioQAAQ} and multimodal contexts~\cite{Kembhavi2017AreYS}. Some of these use questions written by humans~\cite{agrawal2017vqa, Kembhavi2016ADI, Kembhavi2017AreYS, Krishna2016VisualGC} and others use questions that are generated automatically~\cite{Andreas2016NeuralMN, Johnson2016CLEVRAD, Kahou2017FigureQAAA}. \datasetshort\ is set in a photo-realistic simulation environment and uses automatically generated questions. In contrast to the aforementioned datasets that only require the agent to observe the content passively, \datasetshort\ requires the agent to interact with a dynamic environment.

The first deep architectures designed for VQA involved using an RNN to encode the question, using a CNN to encode the image and combining them using fully connected layers to yield the answer~\cite{agrawal2017vqa, malinowski2015ask}. 
More recently, modular networks~\cite{Andreas2016NeuralMN, Hu2017LearningTR, Johnson2017InferringAE} that construct an explicit representation of the reasoning process by exploiting the compositional nature of language have been proposed. Similar architectures have also been applied to the video domain with extensions such as spatiotemporal attention~\cite{Jang2017TGIFQATS, Mun2016MarioQAAQ}. 
Our proposed approach to question answering allows the agent to interact with its environment and is thus fundamentally different to past QA approaches. However, we note that approaches such as visual attention and modularity can easily be combined with our model to provide further improvements.

\noindent \textbf{Reinforcement Learning (RL):} 
RL algorithms have been employed in a wide range of problems including locomotion~\cite{kohl2004policy}, obstacle detection~\cite{michels2005high} and autonomous flight~\cite{kollar2008trajectory, Sadeghi2016CAD2RLRS}. 
Of particular relevance to our approach is the area of hierarchical reinforcement learning (HRL), which consists of a high level controller and one or more low level controllers. The high-level controller selects a subtask to be executed and invokes one of the low level controllers. The advantage of HRL is that it allows the model to operate at multiple levels of temporal abstraction. Early works proposing HRL algorithms include~\cite{Dietterich2000HierarchicalRL, Parr1997ReinforcementLW, Sutton1999BetweenMA}. More recent approaches include~\cite{kulkarni2016hierarchical} who propose hierarchical-DQN with an intrinsically motivated RL algorithm,~\cite{Tessler2017ADH} who use HRL to create a lifelong learning system that has the ability to reuse and transfer knowledge from one task to another, and~\cite{oh2016communicating} who use HRL to enable zero shot task generalization by learning subtask embeddings that capture correspondences between similar subtasks. Our use of HRL primarily lets us learn at multiple time scales and its integration with the semantic memory lets us divide the complex task of \taskshort\ into more concrete tasks of navigation, detection, planning etc. that are easier to train.

RL techniques have also recently been applied to QA tasks, most notably by~\cite{Johnson2017InferringAE} to train a program generator that constructs an explicit representation of the reasoning process to be performed and an execution engine that executes the program to predict the answer. 

\noindent \textbf{Visual Navigation:}
The majority of visual navigation techniques fall into three categories: offline map-based, online map-based, and map-less approaches. Offline map-based techniques~\cite{borenstein1989real, borenstein1991vector, kim1999symbolic,oriolo1995line} require the complete map of the environment to make any decisions about their actions, which limits their use in unseen environments. Online map-based methods~\cite{davison2003real, engel2014lsd, mur2015orb,  sim2006autonomous, tomono20063, wooden2006guide} often construct the map while exploring the environment. The majority of these approaches use the computed map for navigation only, whereas our model constructs a rich semantic map which is used for navigation as well as planning and question answering. 
Map-less approaches~\cite{haddad1998reactive, jaegle2016fast, linegar2016made, saeedi2006vision, zhu2017target} which use techniques such as obstacle avoidance and feature matching, depend upon implicit representations of the world to perform navigation, and lack long-term memory capabilities.
Recently Gupta \etal~\cite{gupta2017cognitive} proposed a joint architecture for a \emph{mapper} that produces a spatial memory and a \emph{planner} that can plan paths. The similarities between our works lie in the usage of a hierarchical system and a spatial memory. In contrast to their work, navigation is not the end goal of our system, but a subtask towards question answering, and our action space is more diverse as it includes interaction and question answering. 

\noindent \textbf{Visual Planning:} 
To answer questions such as \emph{Do I need to buy milk?} an agent needs to plan a sequence of actions to explore and interact with the environment. A large body of research on planning algorithms~\cite {dornhege2009integrating,fikes1971strips,kaelbling2011hierarchical,srivastava2014combined,srivastava2013using} use high-level formal languages. These techniques are designed to handle low-dimensional state spaces but do not scale well to high-dimensional state spaces such as natural images. 

Other relevant work includes visual navigation~\cite{zhu2017target} and visual semantic planning~\cite{zhu2017visual} which both use the AI2-THOR environment \cite{thor}. The former tackles navigation, and the latter focuses on high level planning and assumes an ideal low level task executor; in contrast, our model trains low level and high level controllers jointly. Also, both these approaches do not generalize well to unseen scenes, whereas our experiments show that we do not overfit to previously encountered environments. Finally, these methods lack any sort of explicit map, whereas we construct a semantic map which helps us navigate and answer questions.

Recently Chaplot \etal~\cite{Chaplot2017GatedAttentionAF} and Hill \etal~\cite{Hill2017UnderstandingGL} have proposed models to complete navigation tasks specified via language (e.g. \emph{Go to the red keycard}) and trained their systems in simulated 3D environments. These models show the ability to generalize to unseen instructions of seen concepts. In contrast, we tackle several question types that require a variety of navigation behaviours and interaction, and the environment we use is significantly more photo-realistic. In our experiments, we compare our proposed \modelshort\ model to a baseline system (A3C in Section~\ref{sec:experiments}) that very closely resembles the model architectures proposed in~\cite{Chaplot2017GatedAttentionAF} and~\cite{Hill2017UnderstandingGL}.

\noindent \textbf{Visual Learning by Simulation:}
There has been an increased use of simulated environments and game platforms to train computer vision systems to perform tasks such as learning the dynamics of the world~\cite{mottaghi2016newtonian, mottaghi2016happens, wu2015galileo}, semantic segmentation~\cite{handa2016understanding}, pedestrian detection~\cite{marin2010learning}, pose estimation~\cite{papon2015semantic} and urban driving~\cite{shafaei2016play,chen2015deepdriving,richter2016playing,ros2016synthia}. Several of these are also interactive making them suitable to learn control, including~\cite{bellemare2013arcade,kempka2016vizdoom,thor, lerer2016learning,wymann2000torcs}. We choose to use AI2-THOR~\cite{thor} in our work since it provides a photo-realistic and interactive environment of real world scenes, making it very suitable to train \taskshort\ systems that might be transferable to the real world.
\section{Learning Framework}

\subsection{Actionable Environment}
Training and evaluating interactive agents in the real world is currently prohibitive from the standpoint of operating costs, scale, time, and research reproducibility. A far more viable alternative is to use simulated environments. However, the framework should be visually realistic, allow interactions with objects, and have a detailed model of the physics of the scene so that agent movements and object interactions are properly represented. Hence, we adopt the AI2-THOR environment~\cite{thor} for our purposes. AI2-THOR is a photo-realistic simulation environment of 120 rooms in indoor settings, tightly integrated with a physics engine. Each scene consists of a variety of objects, from furniture such as couches, appliances such as microwaves and smaller objects such as crockery, cutlery, books, fruit, etc. Many of these objects are actionable such as fridges which can be opened, cups which can be picked up and put down, and stoves which can be turned on and off.

\begin{table}
\begin{center}
\begin{tabular}{ |l|l|l|  }
 \hline
 \multicolumn{3}{|c|}{\dataset ~Statistics} \\
 \hline
       & Train &Test\\
 \hline
Existence & 25,600 & 640  \\
Counting & 25,600 & 640  \\
Spatial Relationships & 25,600 & 640 \\
Rooms & 25 & 5 \\
Total scene configurations (s.c.) & 76,800 & 1,920 \\
Avg \# objects per (s.c.) & 46 & 41 \\
Avg \# interactable objects (s.c.) & 21 & 16 \\
Vocabulary Size & 70 & 70 \\
 \hline
\end{tabular}
\end{center}
\caption{This table shows the statistics of our proposed dataset in a variety of question types, objects and scene configurations.}
\label{tab:dataset}
\vspace{-5mm}
\end{table}

\subsection{\dataset}
\datasetshort\ is a question answering dataset built upon AI2-THOR~\cite{thor}. It consists of over 75,000 multiple choice questions for three different question types (table~\ref{tab:dataset} shows more detailed statistics). Each question is accompanied by a scene identifier and a unique arrangement of movable objects in the scene. Figure \ref{fig:teaser} shows three such examples. The wide variety of configurations in \datasetshort\ prevent models from memorizing simple rules like ``apples are always in the fridge'' and render this dataset challenging. \datasetshort\ consists of several question types including: Existence questions (\emph{Is there an apple in the kitchen?}), Counting questions (\emph{How many forks are present in the scene?}),
and Spatial Relationship questions (\emph{Is there lettuce in the fridge? / Is there a cup on the counter-top?}). Questions, ground truth answers, and answer choices are generated automatically. Since natural language understanding is not a focus of this dataset, questions are generated using a set of templates written down a priori. Extending \datasetshort\ to include more diverse questions generated by humans is future work. \datasetshort\ is a balanced dataset that prevents models from obtaining high accuracies by simply exploiting trivial language and scene configuration biases. Similar to past balanced VQA datasets~\cite{Goyal2016MakingTV}, each question is associated with multiple scene configurations that result in different answers to the question. We split the 30 kitchen rooms into 25 train and 5 test, and have 1024 unique (question, scene configuration) pairs for each (room, question type) pair in train, and 128 in test. An episode is finished when the \emph{Answerer} is invoked. We evaluate different methods using Top-1 accuracy.

\subsection{Agent and Objects}
The agent in our environments has a single RGB camera mounted at a fixed height. An agent can perform one of five navigation actions (move ahead 25 cm, rotate 90 degrees left or right, look up or down 30 degrees). We assume a grid-world floor plan that ensures that the agent always moves along the edges of a grid and comes to a stop on a node in this grid. The agent can perform two interaction actions (open and close) to manipulate objects. A wide variety of objects (fridges, cabinets, drawers, microwaves, etc.) can be interacted with. If there are multiple items in the current viewpoint which can be opened or closed, the environment chooses the one nearest to the center of the current image. The success of each action depends on the current state of the environment as well as the agent's current location. For instance, the agent cannot open a cabinet that is more than 1 meter away or is not in view, or is already open, and it cannot walk through a table or a wall.

\section{Model}

We propose \modelshort\ (\model), consisting of a hierarchy of controllers that operate at multiple levels of temporal abstraction and a rich semantic memory that aids in navigation, interaction, and question answering. Figure~\ref{fig:overview} provides an overview of \modelshort. We now describe each of \modelshort's components in greater detail.

\begin{figure}[h]
\begin{center}
\includegraphics[width=1.0\linewidth]{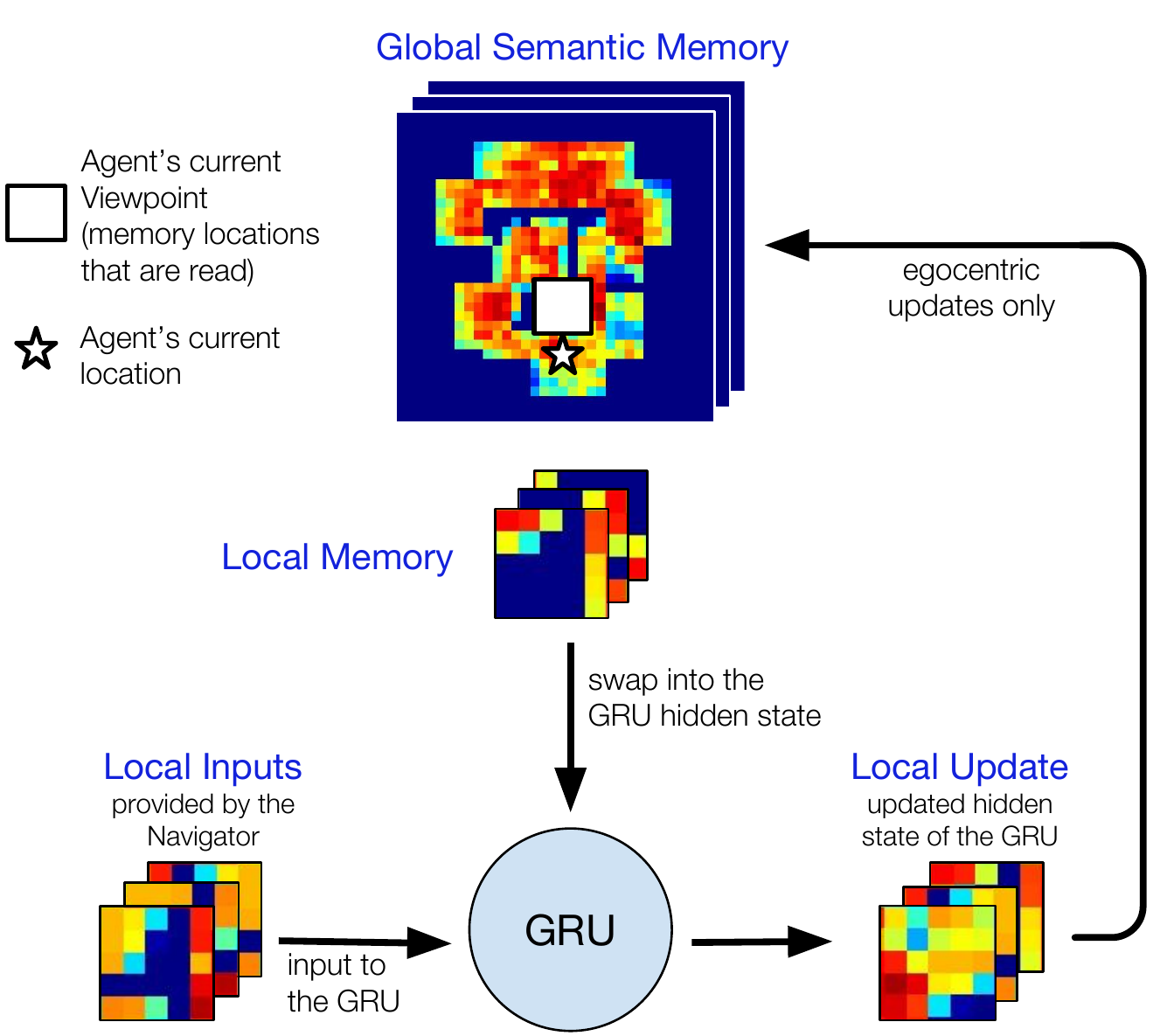}
\caption{An overview of the \gru\ (\grushort): The \grushort\ only allows writing to a local window within the memory, dependent on the agent's current location and viewpoint.} 
\vspace{-7mm}
\label{fig:spatial_gru}
\end{center}
\end{figure}

\subsection{Spatial Memory}
\label{sec:spatial_memory}
Several question types require the agent to keep track of objects that it has seen in the past along with their locations. For complex scenes with several locations and interactable objects, the agent needs to hold this information in memory for a long duration. This motivates the need for an explicit external memory representation that is filled by the agent on the fly and can be accessed at any time. To address this, \modelshort\ uses a rich semantic spatial memory that encodes a semantic representation of each location in the scene. Each location in this memory consists of a feature vector encoding object detection probabilities, free space probability (a 2D occupancy grid), coverage (has the agent inspected this location before), and navigation intent (has the agent attempted to visit this location before). We propose a new recurrent layer formulation: \gru\ (\grushort) to represent this memory, illustrated in Figure~\ref{fig:spatial_gru}. The \grushort\ maintains an external global spatial memory represented as a 3D tensor. At each time step, the \grushort\ swaps in local egocentric copies of this memory into the hidden state of the GRU, performs computations using current inputs, and then swaps out the resulting hidden state into the global memory at the predetermined location. This speeds up computations and prevents corrupting the memory at locations far away from the agent's current viewpoint. When navigating and answering questions, the agent can access the full memory, enabling long-term recall from observations seen hundreds of states prior. Furthermore, only low level controllers have read-write access to this memory. Since the \emph{Planner} only makes high level decisions, without interacting with the world at a lower level, it only has read access to the memory.

\begin{figure}[tp]
\begin{center}
\includegraphics[width=1.0\linewidth]{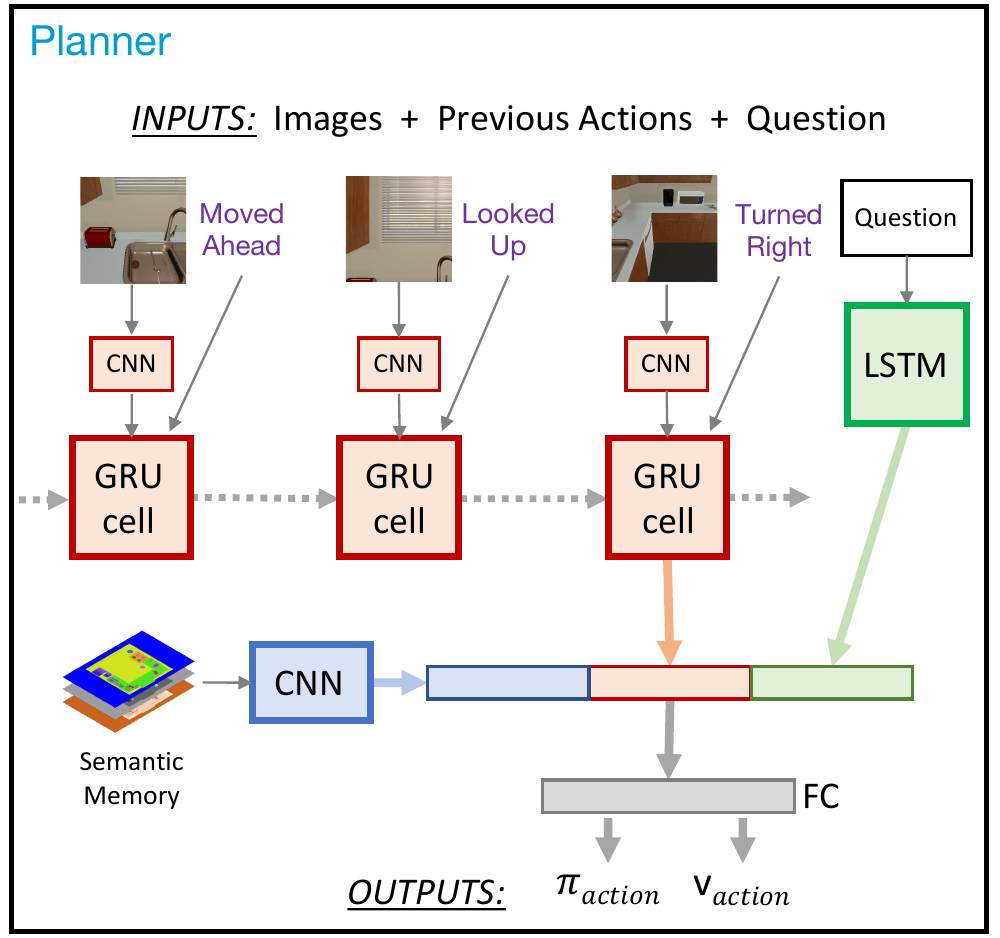}
\caption{Schematic representation of the \emph{Planner}} 
\vspace{-7mm}
\label{fig:planner}
\end{center}
\vspace{-3mm}
\end{figure}

\subsection{Planner}
The high level \emph{Planner} invokes low level controllers in order to explore the environment, gather knowledge needed to answer the given question, and answer the question. We frame this as a reinforcement learning problem where the agent must issue the fewest possible commands that result in a correct answer. The agent must learn to explore relevant areas of the scene based on learned knowledge (e.g. apples are often in the fridge, cabinets are openable, etc.), the current memory state (e.g. the fridge is to the left), current observations (e.g. the fridge is closed) and the question. At every timestep, the \emph{Planner} chooses to either invoke the \emph{Navigator} providing a relative location in a 5x5 grid in front of the agent, invoke the \emph{Scanner} with a direction such as up or left, invoke the \emph{Manipulator} with open/close commands on a nearby object, or invoke the \emph{Answerer} for a total of 32 discrete actions. It does this by producing a policy $\pi$ consisting of probabilities $\pi_i$ for each action, and a value $v$ for the current state. $\pi$ and $v$ are learned using the A3C algorithm \cite{mnih2016asynchronous}. Figure~\ref{fig:planner} shows a schematic of the \emph{Planner}. It consists of a GRU which accepts at each time step the current viewpoint (encoded by a CNN) and the previous action. The \emph{Planner} has read only access to the semantic memory centered around the agent's current location. The output of this GRU is combined with the question embedding and an embedding of the nearby semantic spatial memory to predict $\pi$ and $v$. The agent receives a fixed reward/penalty based on answering correctly/incorrectly. It is also provided a constant time penalty to encourage efficient explorations of the environment and quick answering, as well as a penalty for attempting to perform invalid actions. The agent is also given intermediate rewards for increasing the ``coverage'' of the environment, effectively training the network to maximize the amount of the room it has explored as quickly as possible. Finally, at each time step, the \emph{Planner} also predicts which high level actions are viable given the current world state. In many locations in the scenes, certain navigation destinations are unreachable or there are no objects to interact with. Predicting possible/impossible actions at each time step, allows gradients to propagate through all actions rather than just the chosen action. This leads to higher accuracies and faster convergence (see section~\ref{sec:invalid} for more details).

\subsection{Low level controllers}

\begin{table*}[ht]
\begin{center}
\resizebox{\textwidth}{!}{
\begin{tabular}{ |l||l|l|l|l|l|l|}
 \hline
 & \multicolumn{2}{|c|}{Existence} & \multicolumn{2}{|c|}{Counting} & \multicolumn{2}{|c|}{Spatial Relationships} \\ \hline
 Model & Accuracy & Length & Accuracy & Length & Accuracy & Length \\ \hline
Most Likely Answer Per Q-type (MLA) & 50 & - & 25 & - & 50 & - \\
A3C with ground truth (GT) detections & 48.59 & 332.41 & 24.53 & 998.32 & 49.84 & 578.71 \\ 
HIMN with YOLO~\cite{yolov3} detections & \textbf{68.47} & 318.33 & \textbf{30.43} & 926.11 & \textbf{58.67} & 516.23 \\
Human (small sample) & 90 & 58.40 & 80 & 81.90 & 90 & 43.00 \\ \hline
\end{tabular}
}
\end{center}
\caption{This tables compares the test accuracy and episode lengths of question answering across different models and question types.}
\label{table:accuracy}
\vspace{-3mm}
\end{table*}

\noindent \textbf{Navigator} The \emph{Navigator} is invoked by the \emph{Planner} which also provides it with the relative coordinates of the target location. Given a destination specified by the \emph{Planner} and the current estimate of the room's occupancy grid, the \emph{Navigator} runs A* search to find the shortest path to the goal. As the \emph{Navigator} moves through the environment, it uses the \grushort\ to produce a local (5x5) occupancy grid given the current visual observation. This updates the global occupancy estimate, and prompts a new shortest-path computation. This is a fully supervised problem and can be trained with the standard sigmoid-cross-entropy. The \emph{Navigator} also invokes the \emph{Scanner} to obtain a wide angle view of the environment. Given that the requested destination may be outside the bounds of the room or otherwise impossible (e.g. at a wall or other obstacle), the \emph{Navigator}'s network also predicts a termination signal, and returns control to the \emph{Planner} when the prediction passes a certain threshold.

\noindent \textbf{Scanner} The \emph{Scanner} is a simple controller which captures images by rotating the camera up, down, left, or right while maintaining the agent's current location. The \emph{Scanner} calls the \emph{Detector} on every new image.

\noindent \textbf{Detector} Object detection is a critical component of \modelshort\ given that all questions in \datasetshort\ involve one or more objects in the room. We use YOLOv3~\cite{yolov3} fine-tuned on the AI2-THOR training scenes as an object detector. We estimate the depth of an object using the FRCN depth estimation network~\cite{laina2016deeper} and project the probabilities of the detected objects onto the ground plane. Both of these networks operate at real-time speeds, which is necessary since they are invoked on every new image. The detection probabilities are incorporated into the spatial memory using a moving average update rule. We also perform experiments where we substitute the trained detector and depth estimator with oracle detections. Detections provided by the environment still requires the network to learn affordances. For instance, the network must learn that \emph{microwaves can be opened}, \emph{apples can be in fridges}, etc.

\noindent \textbf{Manipulator} The \emph{Manipulator} is invoked by the \emph{Planner} to manipulate the current state of an object. For example, opening and closing the microwave. This leads to a change in the visual appearance of the scene. If the object is too far away or out of view, the action will fail.

\noindent \textbf{Answerer} The \emph{Answerer} is invoked by the \emph{Planner} to answer the question. It uses the current image, the full spatial memory, and the question embedding vector to predict answer probabilities $a_i$ for each possible answer to the question. The question vector is tiled to create a tensor with the same width and height as the spatial memory. These are depthwise concatenated with the spatial memory and passed through 4 convolution and max pool layers followed by a sum over the spatial layers. This output vector is fed through two fully connected layers and a softmax over possible answer choices. After the \emph{Answerer} is invoked, the episode ends, whether the answer was correct or not.

\subsection{Training}
The full system is trained jointly. However, since the individual tasks of the controllers are mostly independent, we are able to pretrain them separately. Our initial analysis showed that this leads to faster convergence and better accuracy than training end-to-end from scratch. We outline our training procedures below.

\noindent \textbf{Planner:} To pretrain the \emph{Planner}, we assume a perfect \emph{Navigator} and \emph{Detector} by using the ground truth shortest path for the \emph{Navigator} and the ground truth object information for the \emph{Detector}. 

\noindent \textbf{Navigator:} We pretrain the \emph{Navigator} by providing pairs of random starting points and goal locations.

\noindent \textbf{Answerer:} The \emph{Answerer} is pretrained by using ground-truth partial semantic maps which contain enough information to answer the current question correctly.

\noindent \textbf{Detector:} The \emph{Detector} is pretrained by fine-tuning YOLOv3 \cite{yolov3} on the AI2-THOR training scenes. It is trained to identify small object instances which may repeat in multiple scenes (apples, forks, etc.) as well as large object instances which are unique to each scene (e.g. each fridge model will only exist in one scene).

\noindent \textbf{Scanner and Manipulator:} There are no trainable parameters for these controllers in our current setup. Their behavior is predefined by the AI2-THOR environment.

\noindent \textbf{Joint Training} After all trainable controllers are pretrained, we update the model end-to-end.

\begin{table*}
\begin{center}
\resizebox{\textwidth}{!}{
\begin{tabular}{ |l||l|l|l|l|l|l|}
 \hline
 & \multicolumn{2}{|c|}{Existence} & \multicolumn{2}{|c|}{Counting} & \multicolumn{2}{|c|}{Spatial Relationships} \\ \hline
 Model & Accuracy & Length & Accuracy & Length & Accuracy & Length \\ \hline
HIMN with YOLO~\cite{yolov3} detections & 68.47 & 318.33 & 30.43 & 926.11 & 58.67 & 516.23 \\ 
HIMN with GT detection & 86.56 & 679.70 & 35.31 & 604.79 & 70.94 & 311.03 \\
HIMN with GT detection and oracle navigator (HIMN-GT) & \textbf{88.60} & 618.63 & \textbf{48.44} & 871.12 & \textbf{72.50} & 475.55 \\ 
HIMN-GT Question not given to planner & 50.00 & 150.60 & 24.50 & 293.33 & 50.25 & 118.09 \\ 
HIMN-GT No loss on invalid actions & 49.84 & 659.28 & 24.84 & 911.46 & 50.00 & 613.50 \\ \hline
\end{tabular}
}
\end{center}
\vspace{-3mm}
\caption{Ablation experiments on the \modelshort\ model.}
\label{table:ablation}
\vspace{-3mm}
\end{table*}

\section{Experiments}
\label{sec:experiments}

We evaluate \modelshort\ on the \datasetshort\ dataset, using Top-1 question answering accuracy. An initial baseline of Most likely answer per Question-Type (MLA) shows that the dataset is exactly balanced. Additionally, because we construct the data such that each generated question has a scene configuration for each answer possibility, there is no possible language bias in the dataset. The learned baseline that we compare to (A3C), is based on a common architecture for reinforcement learning used in past works including for visual semantic planning~\cite{zhu2017visual} and task oriented language grounding~\cite{Chaplot2017GatedAttentionAF, Hill2017UnderstandingGL}. We extend this for the purpose of question answering. Since \modelshort\ has access to object detections provided by either the environment or YOLO~\cite{yolov3}, we also provide detections to the baseline. For the baseline model, at each time-step, the raw RGB image observed by the agent is concatenated depth wise with object detections (one channel per object class). This tensor is passed through convolutional layers and fed into a GRU. The question is passed through an LSTM. The output of the LSTM and GRU are concatenated, and passed through two fully connected layers to produce probabilities $\pi_i$ for each action and a value $v$. The output of the first fully connected layer is also passed to an answering module that consists of two more fully connected layers with a softmax on the space of all possible answers. The model is trained using the A3C algorithm for action probabilities and a supervised loss on the answers. We also provide a human baseline of random questions on each question type.

Table~\ref{table:accuracy} shows the test accuracies and the average episode lengths for the proposed \modelshort\ model and baselines for each question type. \modelshort\ significantly outperforms the baselines on all question types, both with YOLO object detections as well as ground truth object detections. Surprisingly, the A3C baseline performs slightly worse than random chance even with ground truth detections. We conjecture that this is because there is no explicit signal for when to answer, and no persistence of object detections. The A3C model is not able to associate object detections with the question, and thus has no reason to remember detections for long periods of time. Because \modelshort\ does not overwrite its entire spatial memory at each timestep, object detections persist for much longer, and the \emph{Answerer} can better learn the associations between the questions and the objects. \modelshort\ further benefits from a spatial memory in counting and spatial relationship questions because these require much more spatial reasoning than existence questions. Additionally, because A3C does not learn to answer questions, it also does not learn to efficiently explore the environments, as most of the reward comes from answering questions correctly. \modelshort, on the other hand, traverses much more of the environment, only answering when it is confident that it has sufficiently explored the room. This indicates that \modelshort\ (which uses an explicit semantic spatial memory with egocentric updates) is more effective than A3C (which uses a standard fully-connected GRU) at (a) Estimating when the environment has been sufficiently explored, given the question (b) Keeping track of past observations for much longer durations, which is important in determining answers for questions that require a thorough search of the environment, and (c) Keeping track of multiple object instances in the scene, which may be observed several time steps apart (which is crucial for answering counting questions). 

\subsection{Ablation Analysis}
We perform four ablative experiments on our network structure and inputs, shown in table \ref{table:ablation}. First, we use the ground truth object detections and depth instead of YOLO~\cite{yolov3} and FRCN depth~\cite{laina2016deeper}. This adds a dramatic improvement to our model owing primarily to the fact that without detection mistakes, the \textit{Answerer} can be more accurate and confident. Secondly, we substitute our learned navigation controller with an oracle \emph{Navigator} that takes the shortest path in the environment. When the optimal \emph{Navigator} is provided, \modelshort\ further improves. This is because the \emph{Planner} can more accurately direct the agent through the environment, allowing it to be more efficient and more thorough at exploring the environment. It also takes fewer invalid actions (as seen in table \ref{table:invalid}), indicating that it is less likely to get stuck in parts of the room. In our third ablative experiment, we remove the question vector from the input of the \emph{Planner}, only providing it to the \emph{Answerer}, which results in random performance. This shows that the \emph{Planner} utilizes the question to direct the agent towards different parts of the room to gather information required to answer the question. For instance any question about an object in the fridge requires the planner to know the fridge needs to be opened. If the planner is not told the question, it has no reason to open the fridge, and instead will likely choose to continue exploring the room as exploration often gives more reward than opening an object. Also, some questions can be answered soon after an object is observed (\eg Existence), whereas others require longer explorations (\eg Counting). Having access to the questions can clearly help the \emph{Planner} in these scenarios. Tables~\ref{table:ablation} shows that \modelshort\ does in fact explore the environment for longer durations for Counting questions than for Existence and Spatial Relationship questions. In our final ablation experiment, we remove the loss on invalid actions. If we do not apply any loss on these actions and only propagate gradients through the chosen action, the agent suffers from the difficulty of exploring a large action space and again performs at random chance.

\begin{table}
\begin{center}
\resizebox{\columnwidth}{!}{
\begin{tabular}{ |l|l|l|l|  }
 \hline
 \multicolumn{4}{|c|}{Percentage of invalid actions} \\
 \hline
  \multicolumn{1}{|l|}{Model} & \multicolumn{1}{|c|}{Existence} & \multicolumn{1}{|c|}{Counting} & \multicolumn{1}{|c|}{Spatial Relationships}\\
 \hline
A3C with GT detections & 32.75 & 34.55 & 32.63 \\
HIMN No loss on invalid actions & 56.27 & 53.43 & 51.93 \\
HIMN with YOLO~\cite{yolov3} detections & 6.07 & 5.95 & 6.68 \\ 
HIMN with GT detections & 6.49 & 5.71 & 5.66 \\
HIMN-GT & \textbf{1.79} & \textbf{2.02} & \textbf{1.27} \\
Human & 5.99 & 6.47 & 3.49 \\
 \hline
\end{tabular}
}
\end{center}
\caption{This tables compares the percentage of invalid actions across different models on test. Lower is better.}
\label{table:invalid}
\vspace{-3mm}
\end{table}

\begin{figure*}[t]
\begin{center}
\includegraphics[width=1.0\linewidth]{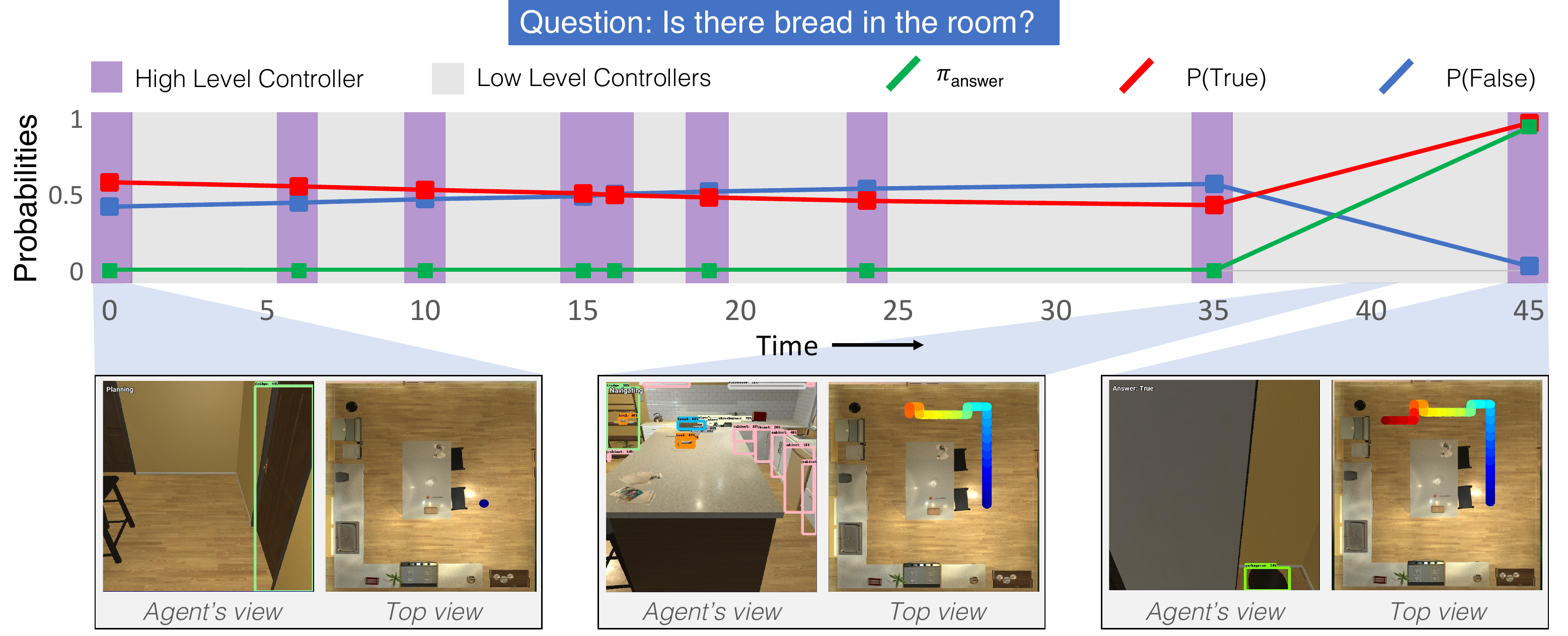}
\caption{A sample trajectory for answering the existence question: \emph{Is there bread in the room?} The purple sections indicate a \emph{Planner} step, and the gray sections indicate a lower level controller such as the \emph{Navigator} is controlling the agent. For a more detailed explanation, refer to section~\ref{sec:qualitative}.} 
\vspace{-8mm}
\label{fig:outputs}
\end{center}
\end{figure*}

\subsection{Invalid Actions}
\label{sec:invalid}
Table~\ref{table:invalid} shows the percentage of invalid actions taken by the different methods. Failed actions are due to navigation failures (failing to see an obstacle) or interaction failures (trying to interact with something too far away or otherwise impossible). 
There is a clear benefit to including a loss on the invalid actions both in terms of QA accuracy, as can be seen in table~\ref{table:ablation}, as well as in terms of percentage of invalid actions performed, shown in table~\ref{table:invalid}. All models in table~\ref{table:invalid} are penalized for every invalid action they attempt, but this only provides feedback on a single action at every timestep. With the addition of a supervised loss on all possible actions, the percentage of invalid actions performed is nearly an order of magnitude lower. By directly training our agent to recognize affordances (valid actions), we are able to mitigate the difficulties posed by a large action space, allowing the \emph{Planner} to learn much more quickly. The validity loss also serves as an auxiliary task which has been shown to aid the convergence of RL algorithms~\cite{jaderberg2016reinforcement}. By replacing the learned \emph{Navigator} with an oracle, we observe that the majority of failed actions are due to navigation failures. We believe that with a smaller step size, we would further reduce the navigation errors at the expense of longer trajectories.

\begin{table}[h]
\begin{center}
\resizebox{\columnwidth}{!}{
\begin{tabular}{|l||l|l|l|l|l|l|}
 \hline
  & \multicolumn{2}{|c|}{Existence} & \multicolumn{2}{|c|}{Counting} & \multicolumn{2}{|c|}{Spatial Relationships} \\ \hline
 Model & S & U & S & U & S & U \\ \hline
HIMN with YOLO~\cite{yolov3} detections & 73.68 & 68.47 & 36.26 & 30.43 & 60.71 & 58.67 \\ 
HIMN with GT detections & 94.00 & 86.56 & 42.38 & 35.31 & 73.38 & 70.94 \\
 \hline
\end{tabular}
}
\end{center}
\caption{This tables compares the accuracy of question answering across different models on Seen (S) and Unseen (U) environments.}
\label{table:train}
\end{table}

\subsection{Generalization in Unseen Environments}
One benefit of \modelshort\ over other RL architectures is that encoding semantic information into a spatial map should generalize well in both seen and unseen environments. Thus, in table \ref{table:train}, we compare \modelshort's performance on seen and unseen environments. Unseen environments tests the agent with questions that occur in 5 never-before-seen rooms, whereas the seen environments use the 25 training rooms but never-before-seen object placements and corresponding questions. Despite our relatively small number of training rooms, table \ref{table:train} shows that our method only loses up to a few percentage points of accuracy when tested on unseen environments. This contrasts with many other end-to-end RL methods which learn deep features that tend to limit their applicability outside a known domain~\cite{zhu2017visual, zhu2017target}. Note that all experiments in previous sections were only performed on unseen environments.

\subsection{Qualitative Results} \label{sec:qualitative}
Figure \ref{fig:outputs} shows a sample run of \modelshort\ for the question ``Is there bread in the room.'' Initially, $P(True)$ and $P(False)$ both start near 50\%. The \emph{Planner} begins searching the room by navigating around the kitchen table. During the initial exploration phase, bread is not detected, and $P(False)$ slowly increases. At timestep 39, the \emph{Navigator} invokes the \emph{Detector}, which sees the bread and incorporates it into the semantic spatial map. However, the \emph{Navigator} does not return control to the \emph{Planner}, as it has not yet reached the desired destination. Upon returning at timestep 45, the \emph{Planner} reads the spatial map, sees the bread, and immediately decides it can answer the question. Thus $\pi_{answer}$ and $P(True)$ both increase to nearly 100\%. For more examples, please see our supplementary video \url{https://youtu.be/pXd3C-1jr98}.

\subsection{Limitations}
Although \modelshort\ performs quite well, it still has several obvious limitations. Due to the 2D nature of the semantic spatial map, \modelshort\ is unable to differentiate between an object being inside a container and being on top of the container. Two obvious extensions of \modelshort\ are storing an explicit height parameter or using multiple 2D slices to construct a 3D map. Secondly, as can be seen in the human experiments in table \ref{table:accuracy}, \modelshort\ is still fairly inefficient at exploring the environment. We plan on investigating more traditional planning algorithms to reduce the time spent exploring previously searched areas. Finally, our templated language model is quite simple, and would not extend to arbitrary questions. We plan on extending {\sc iquad} to include more varied questions, and we will use more expressive language embeddings like \cite{word2vec, glove} in future work.

\section{Conclusion}
In this work, we pose a new problem of Interactive Question Answering for several question types in interactive environments. We propose the \model, consisting of a factorized set of controllers, allowing the  system to learn from long trajectories. We also introduce the \gru\ for updating spatial memory maps. The effectiveness of our proposed model is demonstrated on a new benchmark dataset built upon a high-quality simulation environment for this task. This dataset still presents several challenges to our model and baselines and warrants future research. 

\section{Acknowledgements}
This work was funded in part by the National Science Foundation under contract number NSF-NRI-1637479, NSF-IIS-1338054,  NSF-1652052, ONR N00014-13-1-0720, the Allen Distinguished Investigator Award, and the Allen Institute for Artificial Intelligence. We would like to thank Xun Huang for initial discussions and dataset prototypes. We would also like to thank NVIDIA for generously providing a DGX used for this research via the UW NVIDIA AI Lab (NVAIL). 

\newpage

{\small
\bibliographystyle{ieee}
\bibliography{references}
}

\end{document}


\title{Supplementary Material for IQA: Visual Question Answering in Interactive Environments}
\maketitle

\section{Using Real Object Detection}
\begin{table}[!htbp]
\begin{center}
\resizebox{\columnwidth}{!}{
\begin{tabular}{|l|l|l|l|}
 \hline
 \multicolumn{4}{|c|}{Accuracy of QA} \\
 \hline
       Model & Existence &Counting & Spatial Relationship\\
 \hline
Random & 50 & 25 & 50 \\
A3C with no object detections & 56.9 & 26.42 & 59.1 \\
\textbf{A3C with YOLO object detection} & \textbf{54.29} & \textbf{26.78} & \textbf{55.36} \\
A3C with oracle object detections & 59.5 & 27.1 & 66.2 \\
\textbf{HIMN with YOLO object detection} & \textbf{63.39} & \textbf{35.89} & \textbf{57.14} \\
HIMN with oracle object detection & 69.8 & 32.2 & 65.6 \\
HIMN with oracle object detection and navigator & 73.03 & 45.35 & 71.42 \\
 \hline
\end{tabular}
}
\end{center}
\caption{This tables compares the test accuracy of question answering across different models.}
\label{table:supp_accuracy}
\end{table}

\begin{table}[!htbp]
\begin{center}
\resizebox{\textwidth}{!}{
\begin{tabular}{ |l|l|l|l|l|l|l|l|l|  }
 \hline
 \multicolumn{9}{|c|}{Accuracy of QA Per Answer} \\
 \hline
       Model & Existence & Existence & Counting & Counting & Counting & Counting & Spatial & Spatial \\
       & (N) & (Y) & (0) & (1) & (2) & (3) & Relation (N) & Relation (Y) \\
 \hline
Always Answer Most Likely Value & 57 & 0 & 27 & 0 & 0 & 0 & 52 & 0 \\
A3C with no object detections & 99.69 & 0.83 & 20.81 & 33.04 & 26.53 & 26.32 & 55.02 & 64.94 \\
A3C with YOLO object detection & 84.91 & 14.05 & 41.22 & 35.4 & 42.07 & 44.81 & 39.72 & 42.77 \\
A3C with oracle object detections & 100 & 1.92 & 34.29 & 11.11 & 35.14 & 23.08 & 64.21 & 68.09 \\
HIMN & 92.41 & 40 & 35.14 & 33.93 & 32.17 & 27.52 & 64.44 & 66.67 \\
HIMN with YOLO object detection & 92.14 & 25.62 & 48.32 & 53.57 & 48.30 & 50.66 & 50.93 & 52.89 \\
HIMN with oracle navigator & 100 & 37.6 & 36.24 & 53.57 & 51.02 & 42.11 & 70.82 & 72.29  \\
 \hline
\end{tabular}
}
\end{center}
\caption{Results for each question category broken down by each possible answer.}
\label{table:supp_breakdown}
\end{table}

The modular nature of \modelshort\ allows us to easily swap out and swap in controllers with different architectures. We replace ground truth object sensing (provided by the AI2-THOR framework) with an object detection algorithm (YOLO V2).  We fine-tune YOLO V2 \cite{redmon2016yolo9000} on the AI2-THOR training scenes to show it examples of classes that are not in the MSCOCO dataset such as bread and cabinet. We estimate the depth of an object using the FRCN depth estimation network \cite{laina2016deeper} and project the probabilities of the detected objects onto the ground plane. These detection probabilities are incorporated into the spatial memory using a moving average update rule.

To fairly compare against A3C, we also train the A3C model with YOLO object detection outputs rather than ground truth detections. Table~\ref{table:supp_accuracy} (comparable to Table 2 in the paper) shows the performance of \modelshort\ with YOLO compared to A3C with YOLO. \modelshort\ significantly outperforms this baseline and also outperforms A3C with ground truth object detections on 2 of the 3 question types. \modelshort\ is presumably able to learn robustness to detection noise because it can directly encode the YOLO object probability outputs into the spatial map and incorporate past observations from the same locations much more directly.

We further separate the accuracies for different answers to explore potential biases in the questions as well as in the model behaviors, shown in Table~\ref{table:supp_breakdown}. We find that the questions are mostly balanced, yet Existence is slightly more likely to be false than true. This is a result of filtering out Existence questions where the object is placed in an unobservable location. Because of this, we notice that the A3C models, rather than learn how to explore the environment, simply exploits the bias in the questions. Our model, on the other hand is still quite likely to get true existence questions correct.

\section{Network Architecture}
The full HIMN network can be broken into several networks for navigation, planning, and answering. Their architectures are as follows: \\

\noindent \textbf{Navigation Network} \\
Inputs: 
\begin{itemize}[itemsep=0pt]
    \item Current image at $ 300 \times 300 \times 3 $ resolution
    \item Previous Action One-Hot Vector
    \item Destination
\end{itemize}
Layers:
\begin{itemize}[itemsep=0pt]
    \item Conv: $64 \times 7 \times 7$ kernels, stride 2, ELU activation
    \item Max Pool: $2 \times 2$, stride 2
    \item Conv: $128 \times 5 \times 5$ kernels, stride 1, ELU activation
    \item Max Pool: $2 \times 2$, stride 2
    \item Conv: $256 \times 3 \times 3$ kernels, stride 1, ELU activation
    \item Conv: $256 \times 3 \times 3$ kernels, stride 1, ELU activation
    \item Conv: $256 \times 3 \times 3$ kernels, stride 1, ELU activation
    \item Max Pool: $2 \times 2$, stride 2
    \item Conv: $512 \times 3 \times 3$ kernels, stride 1, ELU activation
    \item Conv: $512 \times 3 \times 3$ kernels, stride 1, ELU activation
    \item Conv: $512 \times 3 \times 3$ kernels, stride 1, ELU activation
    \item Max Pool: $2 \times 2$, stride 2
    \item FC1: Fully Connected on conv output: 1024 units, ELU activation
    \item FC2: Fully Connected on action one-hot: 32 units, ELU activation
    \item FC-Concat: Concatenate (FC1, FC2)
    \item GRU: 1024 units
    \item GRU-Concat: Concatenate(FC-Concat, GRU)
    \item Spatial GRU: $32 \times 5 \times 5$
    \item \textbf{Output} Path Weight: Conv: $1 \times 1 \times 1$, stride 1, activation = $min(max(1, 5 * e^{x}), 200)$
    \item Crop: Spatial GRU at Destination with $5 \times 5$ padding
    \item Conv: $32 \times 3 \times 3$ kernels, stride 1, ELU activation
    \item Conv: $32 \times 3 \times 3$ kernels, stride 1, ELU activation
    \item \textbf{Output} Terminal: Fully Connected, no activation.
\end{itemize}

\noindent \textbf{Planner Network} \\
Inputs: 
\begin{itemize}[itemsep=0pt]
    \item Current image at $ 300 \times 300 \times 3 $ resolution
    \item Semantic map at $RoomW \times RoomH \times NumClasses + 5$
    \item Previous Action One-Hot Vector
    \item Question Encoding
\end{itemize}
Layers:
\begin{itemize}[itemsep=0pt]
    \item Conv: $32 \times 7 \times 7$ kernels, stride 2, ReLU activation
    \item Max Pool: $2 \times 2$, stride 2
    \item Conv: $64 \times 5 \times 5$ kernels, stride 1, ReLU activation
    \item Max Pool: $2 \times 2$, stride 2
    \item Conv: $128 \times 3 \times 3$ kernels, stride 1, ReLU activation
    \item Max Pool: $2 \times 2$, stride 2
    \item Conv: $256 \times 3 \times 3$ kernels, stride 1, ReLU activation
    \item FC1: Fully Connected on conv output: 1024 units, ELU activation
    \item FC2: Fully Connected on action one-hot: 32 units, ELU activation
    \item FC-Concat: Concatenate (FC1, FC2)
    \item GRU: 1024 units
    \item GRU-Concat: Concatenate(FC-Concat, GRU)
    \\
    \item FCQ: Fully connected on Question Encoding: 64 units, ELU activation
    \item Tile FCQ: $RoomW \times RoomH$
    \item Concatenate(Semantic Map, Tile FCQ)
    \item Conv: $64 \times 1 \times 1$ kernels, stride 1, ELU activation
    \item Conv: $64 \times 11 \times 11$ kernels, stride 1, ELU activation
    \item Conv: $128 \times 3 \times 3$ kernels, stride 1, ELU activation
    \item Conv: $256 \times 3 \times 3$ kernels, stride 1, ELU activation
    \item Semantic Features: Conv: $256 \times 3 \times 3$ kernels, stride 1, ELU activation
    \item \textbf{Output}: Viability: Conv: $1 \times 1 \times 1$, stride 1, no activation
    \item Crop: $1 \times 1$ at Semantic Features(current spatial location)
    \item Concatenate (GRU-Concat, Crop)
    \item \textbf{Output}: $V_{action}$: Fully Connected on Concatenate
    \item \textbf{Output}: $\pi_{action}$ Fully connected on Concatenate: 6 units
\end{itemize}

\noindent \textbf{Answerer Network} \\
Inputs: 
\begin{itemize}[itemsep=0pt]
    \item Semantic map at $RoomW \times RoomH \times NumClasses + 5$
    \item Question Encoding
\end{itemize}
Layers:
\begin{itemize}[itemsep=0pt]
    \item Tile Question: $RoomW \times RoomH$
    \item Concatenate: (Semantic Map, Tile Question)
    \item Conv: $64 \times 1 \times 1$ kernels, stride 1, ELU activation
    \item Conv: $128 \times 3 \times 3$ kernels, stride 2, ELU activation
    \item Max Pool: $2 \times 2$, stride 2
    \item Conv: $128 \times 3 \times 3$ kernels, stride 2, ELU activation
    \item Max Pool: $2 \times 2$, stride 2
    \item Conv: $128 \times 3 \times 3$ kernels, stride 2, ELU activation
    \item Max Pool: $2 \times 2$, stride 2
    \item Spatial Sum
    \item Fully Connected: 128 units, ELU activation
    \item Fully Connected: 128 units, ELU activation
    \item \textbf{Output}: Answer: Fully Connected: NumAnswerClasses, no activation
    \item \textbf{Output}: $V_{answer}$: Fully Connected, no activation
    \item \textbf{Output}: $\pi_{answer}$: Fully Connected, no activation
\end{itemize}

\section{Training details}
To train the navigation network, we used the ADAM optimization algorithm with the learning rate of $10^{-4}$, and default Tensorflow constants. We trained with a batch size of 256 for 200,000 iterations. To train the planner and answerer, we use the RMSProp optimization algorithm with a learning rate of $10^{-3}$. We use the learning curriculum described in the paper for 5 million iterations of A3C (5 million distinct interactions with the environment).

{\small
\bibliographystyle{ieee}
\bibliography{00_references}
}